\documentclass{article}

    \PassOptionsToPackage{numbers, compress}{natbib}

     \usepackage[preprint]{neurips_2023}

\usepackage{subfigure} 
\usepackage[utf8]{inputenc} 
\usepackage[T1]{fontenc}    
\usepackage{hyperref}       
\usepackage{url}            
\usepackage{booktabs}       
\usepackage{amsfonts}       
\usepackage{nicefrac}       
\usepackage{microtype}      
\usepackage{xcolor}         
\usepackage{multirow}
\renewcommand{\arraystretch}{1.25}  
\usepackage{rotating} 
\usepackage{listings}
\usepackage{array, cellspace}
\usepackage{float}
\usepackage{soul}

\usepackage{xcolor} 
\usepackage{makecell}

\usepackage{longtable}
\usepackage[misc]{ifsym}

\newcommand{\re}[1]{{\color{black}#1}}

\newcommand{\nameDemo}{Demo Selector}

\title{TPTU-v2: Boosting Task Planning and Tool Usage of Large Language Model-based Agents in Real-world Systems}

\author{%
  Yilun Kong$^\dagger$$^\ddagger$ \\
  kongyilun@sensetime.com
  \And
  Jingqing Ruan$^\dagger$$^\ddagger$ \\
  ruanjingqing@sensetime.com
  \And
  Yihong Chen$^\dagger$$^\ddagger$ \\
  chenyihong@sensetime.com
  \And
  Bin Zhang$^\dagger$$^\ddagger$ \\
  zhangbin11@sensetime.com
  \And
  Tianpeng Bao$^\dagger$ \\
  baotianpeng@sensetime.com
  \And
  Shiwei Shi$^\dagger$ \\
  shishiwei@sensetime.com
  \And
  Guoqing Du$^\dagger$ \\
  duguoqing@sensetime.com
  \And
  Xiaoru Hu$^\dagger$ \\
  huxiaoru@sensetime.com
  \And
  Hangyu Mao$^\dagger$$^\textrm{\Letter}$  \\
  maohangyu@sensetime.com
  \And
  Ziyue Li \\
  zlibn@connect.ust.hk
  \And
  Xingyu Zeng \\
  zengxingyu@sensetime.com
  \And
  Rui Zhao \\
  zhaorui@sensetime.com
  \AND
  \\
  SenseTime Research
}

\begin{document}

\maketitle

\def\thefootnote{$^\dagger$}\footnotetext{These authors contribute equally to this work.}\def\thefootnote{\arabic{footnote}}
\def\thefootnote{$^\ddagger$}\footnotetext{These authors work as research interns at SenseTime Research.}\def\thefootnote{\arabic{footnote}}
\def\thefootnote{$^\textrm{\Letter}$}\footnotetext{The corresponding author.}\def\thefootnote{\arabic{footnote}}

\begin{abstract}
\re{Large Language Models (LLMs) have demonstrated proficiency in addressing tasks that necessitate a combination of task planning and the usage of external tools that require a blend of task planning and the utilization of external tools, such as APIs. 
However, real-world complex systems present three prevalent challenges concerning task planning and tool usage:} (1) The real system usually has a vast array of APIs, so it is impossible to feed the descriptions of all APIs to the prompt of LLMs as the token length is limited; (2) the real system is designed for handling complex tasks, and the base LLMs can hardly plan a correct sub-task order and API-calling order for such tasks; (3) \re{Similar semantics and functionalities among APIs in real systems create challenges for both LLMs and even humans in distinguishing between them.} 
In response, this paper introduces a comprehensive framework aimed at enhancing the Task Planning and Tool Usage (TPTU) abilities of LLM-based agents operating within real-world systems. Our framework comprises three key components designed to address these challenges: (1) the \emph{API Retriever} \re{selects the most pertinent APIs for the user's task among the extensive array available}; (2) \emph{LLM Finetuner} tunes a base LLM so that the finetuned LLM can be more capable for task planning and API calling; (3) the \emph{\nameDemo} adaptively retrieves different demonstrations related to hard-to-distinguish APIs, which is further used for in-context learning to boost the final performance. 
We validate our methods using a real-world commercial system as well as an open-sourced academic dataset, and the outcomes clearly showcase the efficacy of each individual component as well as the integrated framework.
\end{abstract}

\section{Introduction}

Large language models (LLMs) have exhibited remarkable prowess in natural language processing (NLP) \cite{brown2020language,ouyang2022training,gpt4}, encompassing language understanding \cite{devlin2018bert,radford2023robust}, reasoning \cite{wei2022chain,kojima2022large}, and program synthesis \cite{liu2023your,liang2023code}.

However, leveraging LLMs for complex tasks presents formidable challenges. On one hand, LLMs inherently possess limitations in their capabilities. They have been shown to struggle with solving logical problems such as mathematics, and their training data can quickly become outdated as the world evolves. Instructing LLMs to utilize external tools such as calculators, calendars, or search engines can help prevent them from generating inaccurate information and aid them in effectively addressing problems. On the other hand, integrating these models into complex systems transcends mere task understanding. It demands the ability to break down intricate tasks, manipulate various tools, and engage with users in effective interactions. Several research endeavors, known as LLM-based AI Agents \cite{wang2023survey,xi2023rise}, such as AutoGPT \footnote{\url{https://github.com/Significant-Gravitas/Auto-GPT}}, BabyAGI \footnote{\url{https://github.com/yoheinakajima/babyagi}}, and GhatGPT-plugins \footnote{\url{https://openai.com/blog/chatgpt-plugins}}, have made advancements by employing LLMs as central controllers. These endeavors automatically decompose user queries into sub-tasks, execute low-level tool (API) calls for these sub-tasks, and ultimately resolve the overarching problem.





Despite these advances, LLM-based agents still grapple with pressing challenges in real-world applications. Firstly, real-world systems usually have a vast number of APIs, making it impractical to input descriptions of all APIs into the prompt of LLMs due to the token length limitations. Secondly, the real system is designed for handling complex tasks, and the base LLMs often struggle to correctly plan sub-task orders and API-calling sequences for such tasks. Thirdly, the real system is primarily designed around a core purpose, and as a result, certain APIs may overlap and exhibit similar semantics and functionality, creating difficulty in differentiation for both LLMs and humans. How to address these issues could be the critical step for LLM-based Agents towards omniscience and omnipotence in the real world.

In this paper, we propose a framework to improve the \textbf{T}ask \textbf{P}lanning and \textbf{T}ool \textbf{U}sing (TPTU)~\cite{ruan2023tptu,ruan2023tptu-ws} abilities of LLM-based agents in the real-world systems. 
\re{Compare to our TPTU-v1 ~\cite{ruan2023tptu,ruan2023tptu-ws}}, our new framework consists of three key components to address the above three challenges: 
(1) \textbf{API Retriever} recalls the APIs that are most relevant to the user's task from all APIs. The descriptions of these filtered APIs can then be input into LLM as prompts, allowing the LLM to understand and make accurate choices within the filtered API set. (2) \textbf{LLM Finetuner} tunes a base LLM so that the finetuned LLM can be more capable of task planning and API calls, especially for domain-specific tasks. (3) \textbf{\nameDemo} adaptively retrieves different demonstrations related to hard-to-distinguish APIs, which is further used for in-context learning so that LLM can distinguish the subtle differences in the functions and usages of different APIs. Our main contributions can be summarized as follows:

\begin{enumerate}
    \item We identify three practical challenges that LLM-based agents face when it comes to task planning and tool usage in real-world scenarios.
    \item In response to the three challenges mentioned above, we propose an advanced framework composed of three key components: API Retriever, LLM Finetuner, and~\nameDemo.
    \item Extensive experiments in real-world commercial systems demonstrate the effectiveness of each component and the integrated framework, where the tasks are highly complex and closely intertwined with people's lives. We also validate our methods with open-sourced academic datasets.
\end{enumerate}

\section{Methodology}

In response to the typical challenges of deploying LLMs within intricate real-world systems, we propose a comprehensive framework that fundamentally bolsters the capabilities of LLMs in Task Planning and Tool Usage (TPTU). 
This section first introduces our proposed framework, which systemically integrates three specialized components: an API Retriever, an LLM Finetuner, and a~\nameDemo.
Subsequently, we delve into a comprehensive description of each component, elucidating their unique contributions to the overall framework.

\subsection{Framework Overview}


Our comprehensive framework is engineered to enhance the capabilities of LLMs in Task Planning and Tool Usage (TPTU) within complex real-world systems. 
The framework is meticulously designed to address three core challenges: the extensive number of APIs in real-world systems, the complexity of correct task and API call sequencing, and the difficulty in distinguishing between APIs with overlapping functionalities.

\begin{figure}[ht!]
\centering
\includegraphics[width=1\linewidth]{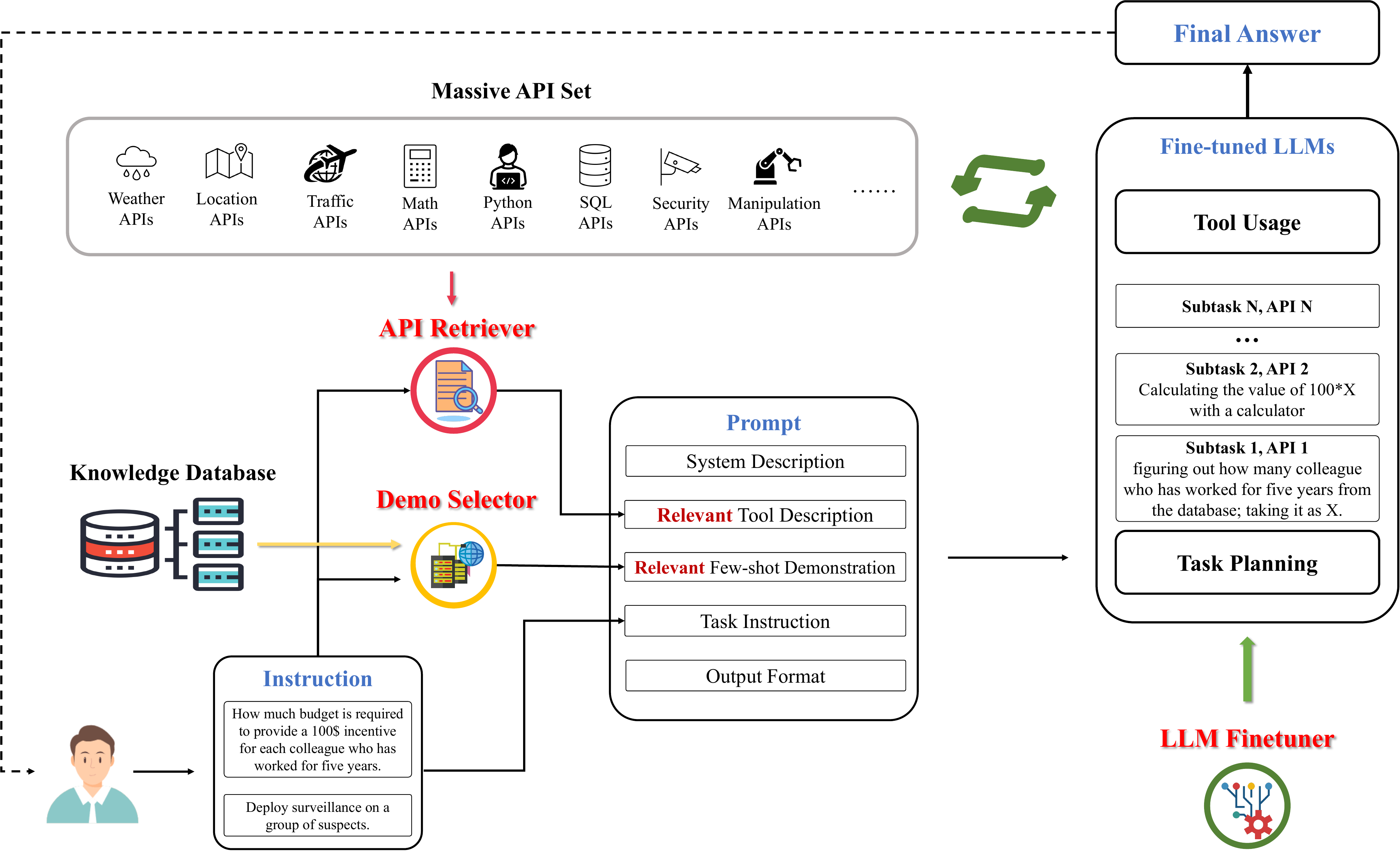}
\caption{The proposed framework.}
\label{fig:main_framework}
\end{figure}

The framework is composed of three pivotal components, depicted in Figure~\ref{fig:main_framework}. 

\begin{enumerate}
    \item \textbf{API Retriever}: This component navigates through an extensive array of APIs to retrieve the most relevant ones based on the user's task. It employs an advanced embedding search technique to understand the semantics of the task and match it with the correct APIs, leveraging a rich Knowledge Database and an API Collection to ensure relevance and accuracy.

    \item \textbf{LLM Finetuner}: This subsystem fine-tunes a base LLM with a meticulously curated dataset, enhancing the model's ability to plan tasks and execute API calls efficiently. The fine-tuning process is informed by diverse datasets, including ones specifically created to increase prompt diversity and address both single-step and multi-step API interactions.

    \item \textbf{\nameDemo}: The~\nameDemo~dynamically retrieves demonstrations related to hard-to-distinguish APIs, facilitating in-context learning for the LLM. This allows the model to discern subtle functional differences between APIs, crucial for generating precise outputs, especially when dealing with similar APIs.
\end{enumerate}

\subsection{API Retriever}

\begin{figure}[ht!]
\centering
\includegraphics[width=1.0\linewidth]{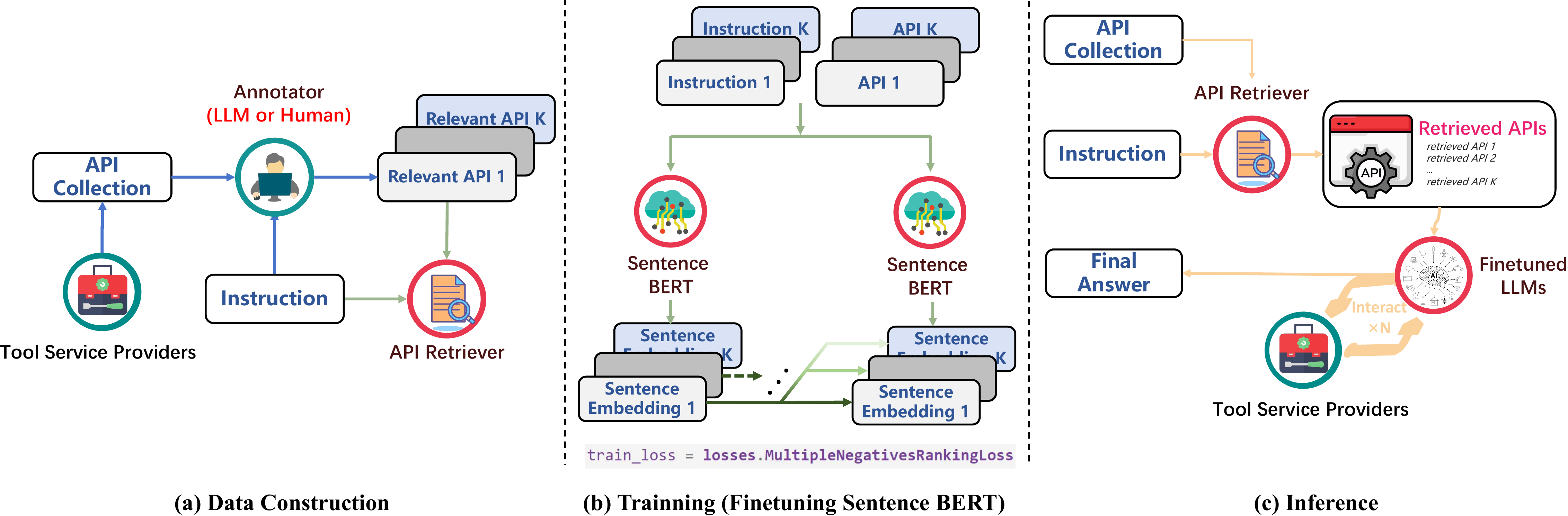}
\caption{The proposed framework of API Retriever.}
\label{fig:api_overview}
\end{figure}

In real-world systems, there exists a massive number of APIs for problem-solving, which poses a severe challenge for the integration of LLMs.
On the one hand, the token limitations inherent to LLMs impede the inclusion of all API descriptions in the model's prompt, potentially surpassing the maximum token length. 
On the other hand, even when the inclusion of numerous APIs does not breach these token constraints, the presence of excessive, task-irrelevant API information can interfere with the model's capacity for accurate planning and answer generation, thereby hindering its operational efficiency. To surmount these challenges, we have developed a novel model explicitly trained to select the APIs of utmost relevance to the task at hand, shown in Figure~\ref{fig:api_overview}.
Building on the overview of the API Retriever framework, we will now give a detailed description of the data collection, training, and inference process.


\subsubsection{Data Collection}

\begin{figure}[ht!]
\centering
\includegraphics[width=1.0\linewidth]{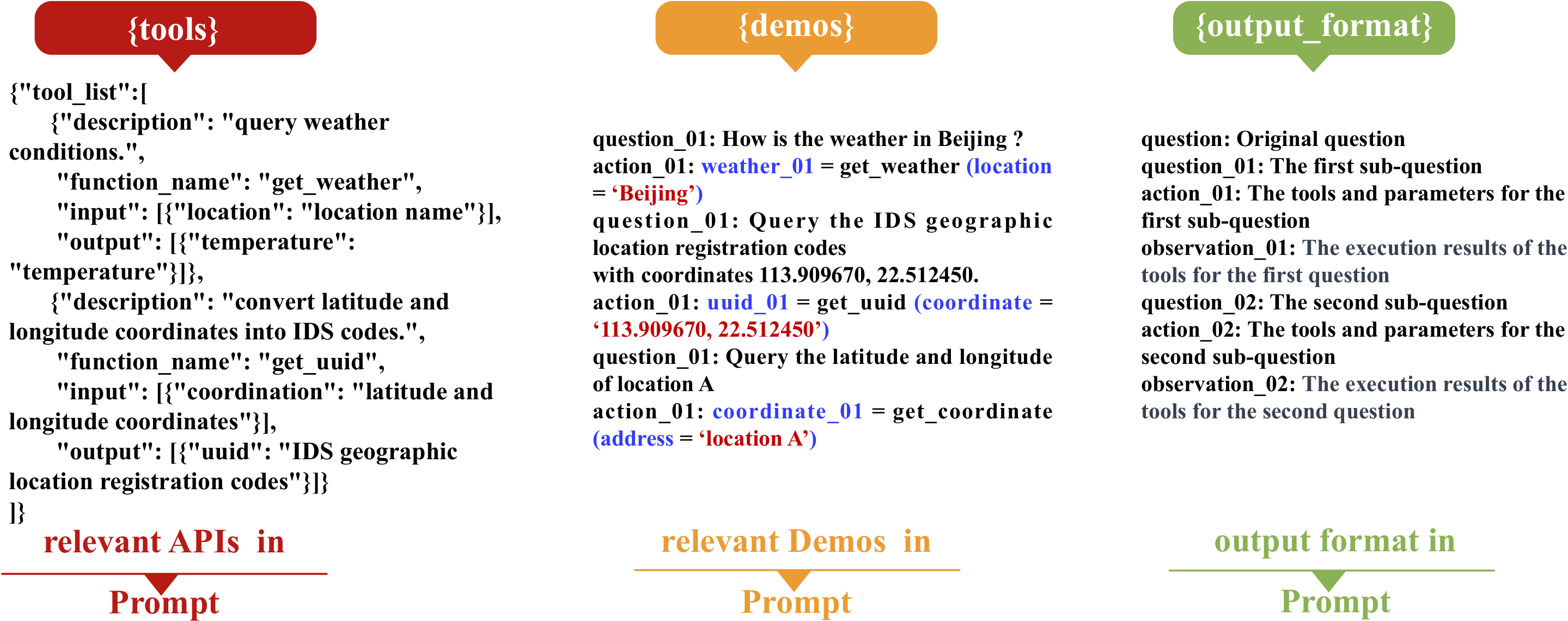}
\caption{The detailed demonstration of the dataset for the API Retriever.}
\label{fig:api_demo}
\end{figure}

The foundation of the API Retriever's effectiveness lies in a rigorous data collection process. 
First, we have collected a comprehensive set of APIs provided by a multitude of external tool services. 
This collection forms the substrate upon which our model is trained. 
To ensure that our system understands the relevance of different APIs to various user queries (instructions), we have instituted a particular annotation process. In this process, human experts, or LLMs, analyze complex user instructions (or tasks) and identify the APIs that are necessary for resolving these instructions. 
This hybrid approach not only enriches our dataset with human expertise but also benefits from the scale and efficiency of LLMs in processing large quantities of data. 
By combining the precision of human annotations with the breadth of LLMs' processing abilities, we create a dataset that is both qualitatively rich and quantitatively vast, laying a solid foundation for the subsequent training phase of the API Retriever.
We give a detailed demonstration of the dataset in Figure~\ref{fig:api_demo}.

\subsubsection{Training}

Following the collection of this annotated data, the training of the API Retriever is conducted to maximize the relevance of the retrieved APIs to the task instruction of users.
The training framework for the API Retriever is depicted as a dual-stream architecture employing Sentence-BERT~\cite{reimers2019sentence}, a variant of the BERT~\cite{devlin2018bert} model optimized for generating sentence embeddings. The training process utilizes pairs of instructions and their corresponding APIs, which are denoted as \textit{Instruction 1} through \textit{Instruction K} and \textit{API 1} through \textit{API K}, respectively.

Each instruction and API description is processed through its own Sentence-BERT model to obtain semantically rich embeddings. This means that for each instruction-API pair, we generate two separate embeddings that encapsulate the semantic essence of the text. The embeddings for the instructions are labeled as \textit{Sentence Embedding 1} to \textit{Sentence Embedding K}, and similarly, the embeddings for the APIs follow the same notation.

The framework employs a training objective known as the Multiple Negatives Ranking Loss \footnote{\url{https://www.sbert.net/docs/package_reference/losses.html\#multiplenegativesrankingloss}}~\cite{henderson2017efficient}. This loss function is designed to contrast a positive pair (a correct association between instruction and an API) against multiple negative pairs (incorrect associations). The goal is to minimize the distance between the embeddings of correct instruction-API pairs while maximizing the distance between the embeddings of incorrect pairs. This goal can formulated as follows.
\begin{equation}
    \mathcal{L} = - \frac{1}{K}\sum\limits_{i = 1}^K {\log \frac{{{e^{sim({s_i},s_i^ + )}}}}{{{e^{sim({s_i},s_i^ + )}} + \sum\nolimits_{j \ne i} {{e^{sim({s_i},s_j^{-} )}}} }}},
\end{equation} 
where $s_i$ and $s_i^+$ denote the \textit{Sentence Embedding i} and the corresponding \textit{Sentence Embedding i} of the API, respectively. $sim(\cdot)$ is the similarity function that calculates the similarity between two vectors (embeddings in this context). 
Our choice for $sim(\cdot)$ is the cosine similarity, which measures the cosine of the angle between two vectors $u$ and $v$, defined as follows.
\begin{equation}
    sim(u, v) = \frac{{u \cdot v}}{{||u||||v||}},
\end{equation}
where $u \cdot v$ is the dot product of vectors, and $||\cdot||$ denotes Euclidean norms (or magnitudes) of the vectors.

During training, this encourages the model to learn a representation space where instructions and their relevant APIs are closer to each other, thus facilitating more accurate retrieval of APIs in response to new instructions.

In summary, the Sentence-BERT models in this framework are fine-tuned to learn the semantic relationships between user instructions and APIs, enabling the API Retriever to discern and prioritize the most relevant APIs for a given task based on their learned embeddings.

\subsubsection{Inference}

The inference diagram illustrates the process that integrates the API Retriever and LLMs with the objective of generating a final answer to a given instruction.

The process commences with an \textit{Instruction}: a user's query or task that needs to be addressed. This \textit{Instruction} is fed into the API Retriever, a component that has been meticulously trained to recognize and select the most relevant APIs from an extensive API Collection. The API Retriever evaluates the instruction, determines the relevant APIs needed to fulfill the task, and retrieves a subset of APIs, denoted as \textit{retrieved API 1} to \textit{retrieved API K}.

Once the relevant APIs are retrieved, they are fed into the tool-level prompt for LLMs to select the accurate APIs to solve certain instructions. It is important to note that there might be multiple interactions (``Interact $\times$ N'') between the LLMs and the Tool Service Providers, which are the actual endpoints of the APIs, indicating that the LLMs may call multiple APIs multiple times to gather the information needed.

Finally, after the LLMs have interacted with the tool service providers as required, they summarize the information gathered from the APIs to construct a ``Final Answer''. This answer is expected to be a comprehensive response to the original instruction, showcasing the system's ability to understand, retrieve, and apply relevant information to solve complex, real-world problems.

\subsection{LLM Finetuner}\label{sec:Finetuner}

While open-sourced LLMs possess strong capabilities, they often encounter limitations due to a lack of specificity and adaptability within complex, specialized, real-world domains. 
Furthermore, certain models may fall short in their generative abilities, struggling to yield high-quality outputs when tasked with challenges. 
To address these issues, we shift our approach from pioneering new fine-tuning methods to \emph{concentrating on the development of a dataset, expressly curated to enhance the fine-tuning process for real-world systems}.
In this context, we will also share some insights during the fine-tuning procedure, providing a clearer understanding of its influence on model performance.

Building upon the foundation established by the introduction, we delve into the fine-tuning of our LLMs using the prevalent method known as Supervised Fine-Tuning (SFT). This mainstream approach to fine-tuning involves adjusting the pre-trained weights of an LLM on a dataset that is labeled with the correct outputs for given inputs. SFT is particularly effective for enhancing model performance in specific domains or tasks, as it steers the model toward the desired output using the provided supervisory signals.

For our fine-tuning process, we have constructed and analyzed three distinct datasets, each representing a unique fine-tuning paradigm:

\begin{enumerate}
    \item \textbf{Training Set v1}: Born out of a need for datasets that accurately mirror real-world scenarios, this initial dataset was constructed by carefully selecting genuine cases, eliminating ineffective data and duplicate cases. Its motivation lies in grounding the SFT in reality, aligning the LLM's understanding with the true data distribution found in practical real-world use. The dataset serves as a preliminary step towards tuning the LLM to adapt to real-world data distribution.

    \item \textbf{Training Set v2}: This dataset is selectively compiled based on prompt functionality, encompassing a total of 745 entries. It is augmented with system-level prompts that include a comprehensive list of features and their descriptions. These enriched prompts serve to provide the LLM with a more detailed understanding of each API's capabilities and constraints. By incorporating a detailed functionality list and descriptions within the prompts, we aim to enhance the model's ability to generate responses that not only match the input query semantically but also align closely with the functional scope of the available APIs. This structured approach to prompt design is crucial for enabling the LLM to navigate the API space with greater precision, particularly when dealing with complex, multi-faceted user requests.

    \item \textbf{Training Set v3}: Recognizing the limitations of our previous dataset, which predominantly featured single-step API calls and suffered from a lack of prompt diversity, we sought to more closely cover real-world scenarios. Training Set v3 was thus meticulously engineered to bridge this domain gap, comprising 660 question-and-answer pairs that reflect the complexity of actual use cases. (1) For prompt diversity, we employ various data augmentation on prompts, e.g., randomly shuffling API orders and adding irrelevant APIs, thus decreasing the risk of over-fitting and enhancing the robustness of the LLM. (2) For instruction diversity, we replace the original user instruction with similar-meaning instructions by means like rewriting-by-LLMs, synonym substitution, and loop-back translation. This makes LLMs more robust to different user instructions during inference. (3) For output diversity, set v3 intentionally includes a balanced mix of 390 single-step API interactions, which solidify the foundational understanding of API functionalities, and an additional 270 multi-step API calls, which introduce the LLM to more complex sequences of operations that are commonly encountered in practice. 
\end{enumerate}



Each dataset is intended to incrementally refine the LLM's ability to parse user inputs, understand the context, and generate precise API calls. 
Finetuning LLMs on these datasets can enhance the ability of LLMs to solve specific real-world tasks.
The analysis of model performance across these datasets provides valuable insights into the effects of prompt diversity and task complexity on the LLM's fine-tuning efficiency and its eventual real-world applicability. 
By systematically evaluating the model's output against these varied fine-tuning paradigms, we enhance its competency in delivering high-quality, contextually appropriate responses in the domain of API interaction.

The insights obtained from the iterative development of these datasets demonstrate the critical importance of dataset quality and construction in the fine-tuning process. 
With each successive version, we observed measurable improvements in the LLM's performance, underscoring the direct impact that well-constructed training data has on the model's ability to handle real-world tasks. 
It is not merely the quantity of data but the relevance, cleanliness, and alignment with actual usage patterns that drive the efficacy of fine-tuning, leading to models that are not only more versatile but also more reliable when deployed in complex real-world applications.




\subsection{\nameDemo}

\begin{figure}[ht!]
\centering
\includegraphics[width=0.75\linewidth]{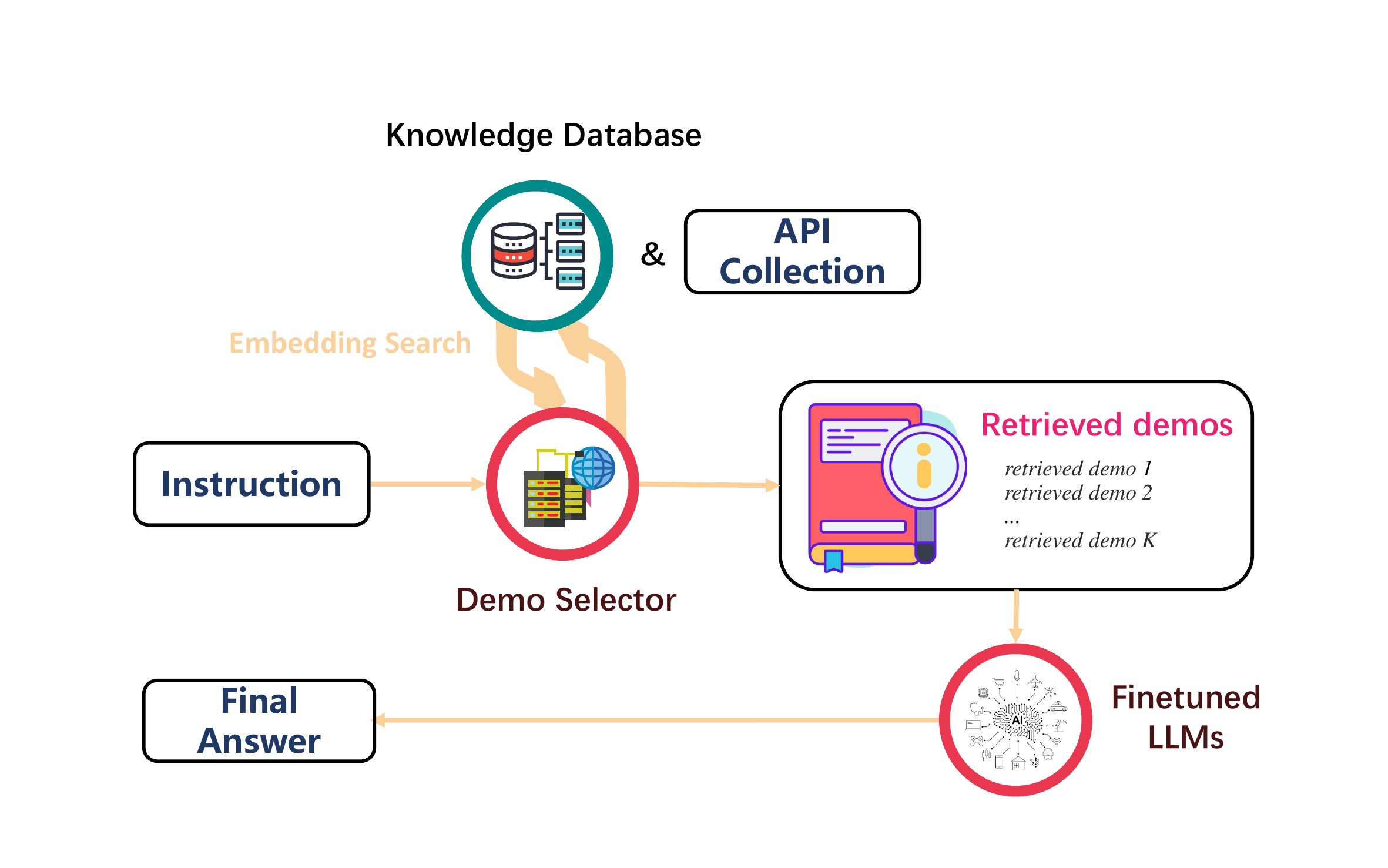}
\caption{The proposed framework of the~\nameDemo.}
\label{fig:demo_framework}
\end{figure}

The~\nameDemo~framework, as shown in Figure~\ref{fig:demo_framework}, plays a crucial role in enhancing the ability of finetuned LLMs to differentiate between APIs with similar functionalities and semantics \footnote{The APIs may have similar semantics and functionality because (1) the real system is primarily designed around a core purpose, so some APIs are relevant; (2) when API retriever is used, the retrieved APIs could be more semantically similar.}. 
Usually, the quality of demonstrations has a very positive influence on promoting the ability of LLMs to disassemble complex tasks.
Here is a detailed description of the main workflow and functionality of the~\nameDemo, guided by the provided knowledge and the information depicted in Figure~\ref{fig:demo_framework}.

The~\nameDemo~is engineered to dynamically retrieve various demonstrations pertinent to APIs that are challenging to distinguish due to their overlapping features. The main workflow begins with an ``Instruction'', which represents a user's query or command that necessitates the utilization of one or more APIs.

Upon receiving an instruction, the~\nameDemo~interacts with two critical resources: the ``Knowledge Database'' and the ``API Collection''. The Knowledge Database contains structured information that could include API documentation, usage examples, and other relevant data that aids in understanding the context and details of each API. The API Collection, on the other hand, comprises the actual API endpoints and their associated metadata.


Then, an embedding searching process is employed to facilitate the retrieval of relevant demonstrations (demos) for a given user query. 
\begin{enumerate}
    \item \textbf{Embedding Generation}. Initially, the user's query $Q$ and demos from the knowledge database $D$ are transformed into vector representations, known as embeddings. Let $emb(Q)$ denote the embedding of the user query, and $emb(D_i)$ represent the embedding of the $i$-th demo in the database, where $i$ ranges from 1 to the total number of examples $N$. Here, we use Sentence-Bert~\cite{reimers2019sentence} as the tool to generate embeddings.
    
    \item \textbf{Similarity Thresholding}. We define a similarity threshold  $\Delta$ to determine the relevance of each demo. The similarity measure  $sim(emb(Q), emb(D_i))$ is computed between the query embedding and each example embedding. This similarity could be calculated using cosine similarity as $sim(emb(Q), emb(D_i)) = \frac{emb(Q) \cdot emb(D_i)}{\|emb(Q)\| \|emb(D_i)\|}$, where $\cdot$ denotes the dot product of the two embeddings, and $\| \cdot \|$ represents the L2 norm.
    
    \item \textbf{Top-k Demo Retrieval}. If the similarity measure for any example exceeds the threshold $sim(emb(Q), emb(D_i)) > \Delta $, we proceed to select the top-k most similar demos $ \{D_{top_1}, D_{top_2}, ..., D_{top_k}\} $ based on their similarity scores. These are regarded as subtask-level demos as they are closely related to the specific task at hand.
    
    \item \textbf{Fallback to API-Level Demos}: In cases where no example exceeds the similarity threshold $ \forall i, sim(emb(Q), emb(D_i)) \leq \Delta $, the process defaults to retrieving demos from the API collection. This involves searching for relevant API-level demos that are aligned with the broader context of the query rather than specific subtask details.
    
\end{enumerate}

The core functionality of the ~\nameDemo~ lies in its adaptability and precision in identifying the most relevant demonstrations for a given task query, ensuring that the LLM is provided with the most contextually appropriate examples for its operation. This process seamlessly prioritizes the retrieval of subtask-level demos that are highly relevant when available, but it can also efficiently fall back on more generalized API-level demos when specific examples do not meet the similarity threshold. By sifting through embeddings and discerning the nuanced differences in API functionalities, the ~\nameDemo~ is capable of selecting from a range of demonstrations, labeled as \textit{retrieved demo 1} to \textit{retrieved demo K}. These context-rich examples are instrumental in illustrating how similar APIs can be distinctively applied, thereby significantly enhancing the LLM's performance in executing complex tasks.

Finally, the interaction between the~\nameDemo~and the finetuned LLMs leads to the generation of a final answer, which is the LLMs' response to the original instruction, informed by the nuanced understanding gained from the demonstrations.


\section{Experiments}

In this section, we present an experiment designed to rigorously evaluate the efficacy of our proposed framework, with a particular focus on the API Retriever, the LLM Finetuner, and the~\nameDemo~components. Our experimental methodology is structured to test the system's performance in a real-world context and an open-source challenge.

We begin by detailing the experimental setup, including the datasets employed. 
This is followed by a series of experiments that systematically assess each component's contribution to the overall functionality of the system. Through a combination of quantitative and qualitative analyses, we aim to demonstrate not only the performance improvements our system achieves over existing approaches but also the specific capabilities it brings to complex task planning and API interaction scenarios.

\subsection{Datasets}

\paragraph{Anonymous Real-world Scenario.}
Diverging from the current scholarly focus on studying the ability to choose the right APIs from a plethora of APIs encompassing various functionalities, in real-world systems, more common and challenging problems often revolve around a few core purposes. It entails choosing the most suitable API from a few dozen APIs, which are closely related in semantics but differ in usage, such as required parameters. Therefore, we constructed a specialized dataset that is composed of 45 APIs revolving around 11 core functionalities, based on a real commercial security system. Note that despite the total number of APIs being only 45, real-world tasks involve different planning trajectories of APIs and their parameters. For example, some trajectories can involve 9 APIs, and the average length of API trajectories is 3.5, which is longer than many open-source datasets \cite{qin2023toolllm,tang2023toolalpaca,li2023api}. The training dataset has been described in Section \ref{sec:Finetuner}. As for the testing dataset, we collected 100 questions for evaluation. Although the number of testing questions is not large, the quality is high. Our product-side colleagues assisted us in collecting this data, including simple questions with fewer than 10 words, as well as challenging questions with more than 100 words. The careful selection of testing questions ensures that they accurately reflect real-world usage scenarios.


\paragraph{Open-source Scenario.}
To ensure the generalizability of our approach across a broader spectrum of tasks and its capability to select appropriate APIs from a myriad of options, we also perform experiments on an open-source dataset, ToolBench\cite{qin2023toolllm}, which contains 16000+ real-world APIs spanning 49 application categories. Besides the variety and quantity of APIs, it is also well conducted with both single-tool and multi-tool scenarios, as well as several multi-step reasoning traces for each query. Thus, ToolBench can simulate a real-world system, and experiments on this dataset can further demonstrate the performance of our framework in complex real-world tasks and its generalization ability across different scenarios. In order to manage the evaluation cost-effectively, we employed a random sampling approach to select 10,000 questions from ToolBench. These questions were then split into three datasets: training, validation, and testing, using a ratio of 7:1:2 respectively. This division allows us to train and fine-tune our models on a substantial amount of data while reserving a separate portion for thorough validation and reliable testing.

\subsection{Experiment on Real-world Scenario}



In our anonymous real-world scenario, we conduct tests to evaluate the effectiveness of the proposed modules in our framework. We begin by assessing the capability of the API retriever on our dataset, achieving a Recall@5 of 84.64\% and Recall@10 of 98.47\% in Table~\ref{tab:exp_api_retr}. These results verify the effectiveness of our method, demonstrating a high level of precision in retrieving relevant APIs, which is crucial for the subsequent task execution phase.

\renewcommand{\arraystretch}{1.4}
\begin{table}[ht!]
\caption{The results of API Retriever on Real-world Scenario}
\begin{center}
\begin{tabular}{ccc}
\hline 
Approaches & Recall@5 & Recall@10 \\
\hline
 API Retriever & 84.64\% & 98.47\% \\
\hline
\\
\end{tabular}
\end{center}
\label{tab:exp_api_retr}
\end{table}

\renewcommand{\arraystretch}{1.4}
\begin{table}[ht!]
\caption{Performance comparison on Real-world Scenario}
\begin{center}
\begin{tabular}{cc}
\hline 
Approaches & Execution Accuracy  \\
\hline
base LLM (no demos and oracle APIs) & 38.89\% \\
base LLM (no demos and oracle APIs)  + API retriever & 43.33\% \\
base LLM (no demos and oracle APIs)  + Demo selector & \ul{95.55\%} \\
finetuned LLM + API retriever & 80\% \\
finetuned LLM + API retriever + Demo selector& \textbf{96.67\%} \\
\hline
\\
\end{tabular}
\end{center}
\label{tab:perf_real}
\end{table}

Moving to the task execution tests, the results are presented in Table~\ref{tab:perf_real}. 
We choose \re{\href{https://internlm.intern-ai.org.cn/}{InternLM} \cite{2023internlm}}, a sophisticated language model developed by Shanghai AI Lab, as our evaluated LLM. The term ``base LLM'' refers to the execution of prompts that do not include demonstrations and utilize the smallest set of Oracle APIs, meticulously selected by human experts.
Intuitively, one might assume that manually selected Oracle APIs would outperform the results obtained using our API Retriever. 
However, contrary to this expectation, our method yields comparable performance. This observation can be attributed to the significant influence of the API order in the prompt on the decisions made by the Language Model (LLM). The relative positioning of APIs within the prompt can have a substantial impact on the LLM's understanding and subsequent decision-making process. The order in which APIs are presented can affect the LLM's interpretation of the context and the relationships between different APIs, ultimately influencing its output. This phenomenon has been previously corroborated by experimental findings in the literature~\cite{lu2021fantastically}. 
Furthermore, in complex scenarios, relying solely on human expertise for precise API selection can be inadequate. It might be a promising approach to automatically retrieve the appropriate API sets.

Regarding the benefits of fine-tuning, the data clearly demonstrates its advantages. The finetuned LLM combined with the API Retriever achieves an 80\% execution accuracy, significantly higher than the base LLM’s performance. This improvement can be attributed to the fine-tuning process, which tailors the LLM more closely to the specifics of the real-world task. It enhances the model's understanding of the context, leading to more accurate and contextually appropriate API calls.                                   

The highest performance is observed when combining the finetuned LLM with both the API Retriever and the Demo Selector, achieving an impressive 96.67\% execution accuracy. This result underscores the effect of integrating fine-tuning with our sophisticated API retrieval and demonstration selection mechanisms. The Demo Selector, in particular, seems to have a substantial impact, likely due to its ability to provide context-rich examples that guide the LLM in making more informed decisions, especially in scenarios involving similar or complex APIs.

In conclusion, our experiments in a real-world setting validate the efficacy of our proposed framework, highlighting the importance of each component and the added value of fine-tuning in enhancing LLM performance for practical applications.

\subsection{Experiment on Open-source Scenario}




In the open-source scenario, we tailor our evaluation to focus primarily on the impact of fine-tuning and the API Retriever, considering that building demonstrations for this context do not significantly contribute to addressing real-world problems. Therefore, the assessment of the Demo Selector is omitted in this scenario.

Initially, we have trained the API Retriever specifically for this scenario, achieving a recall rate of 76.9\%. However, due to the relatively massive nature and high similarity of APIs in this open-source environment, the recall is not as high as expected, which poses a challenge for subsequent performance evaluations.

\renewcommand{\arraystretch}{1.4}
\begin{table}[ht!]
\caption{Performance comparison on Open-source Scenario}
\begin{center}
\begin{tabular}{cc}
\hline 
Approaches & Execution Accuracy  \\
\hline
base LLM & 76.67\% \\
base LLM + API retriever & 53.3\% \\
finetuned LLM + API retriever & 86.7\% \\
\hline
\\
\end{tabular}
\end{center}
\label{tab:perf_open}
\end{table}

As shown in Table~\ref{tab:perf_open}, the execution accuracy of the base LLM stands at 76.67\%. Interestingly, the introduction of the API Retriever results in decreased performance, dropping to 53.3\%. This decline is attributable to several factors. First, the low recall of the API Retriever introduces cumulative errors in the decision-making process. In environments where APIs are relatively massive and highly similar, the increasing complexity of the API Retriever may not align well with task requirements, potentially leading to less optimal API selections. Second, if the API Retriever is trained on a dataset that does not adequately represent the diversity of the open-source scenario, it leads to overfitting. As a result, the API Retriever performs well on training data but poorly generalizes to the broader range of real-world tasks in the evaluation.

Upon implementing fine-tuning in this scenario, an enhancement in performance is observed, with the finetuned LLM combined with the API Retriever reaching an execution accuracy of 86.7\%. This improvement underscores the effectiveness of fine-tuning in adapting the LLM to the specific characteristics and challenges of the open-source environment. The fine-tuning process likely helps the model better understand the nuances of the available APIs and how they correlate with different tasks, resulting in more accurate API calls and decision-making. 

In summary, the open-source scenario highlights the nuanced impacts of our framework's components. It reveals the importance of aligning the capabilities of tools like the API Retriever with the specific demands of the environment and demonstrates the substantial benefits that fine-tuning brings in enhancing model performance in a less complex API ecosystem.

\section{Related Work}

The remarkable capacity for using tools has facilitated the transcendence of human innate physical and cognitive limitations, enhancing our ability to comprehend, plan, and address complex tasks. In turn, the human aptitude for understanding and planning tasks contributes to the judicious selection and usage of appropriate tools. Recently, the swift evolution of LLM has rendered it viable to employ specialized tools and decompose intricate tasks like humans, which inspired significant potential in addressing real-world tasks. Substantial research has been proposed to investigate task planning and tool usage based on LLM separately, however, research that combines these abilities to mutually enhance each other is relatively scarce. TPTU\cite{ruan2023tptu} proposes a complete framework that enhances the agent's ability in task planning and tool utilization for addressing complex tasks.
AgentTuning\cite{zeng2023agenttuning} comprehensively considers various capabilities of LLM, not only task planning and tool usage, enhancing the generalized agent capabilities of open-source LLMs themselves while ensuring their general capabilities are not compromised. Some excellent reviews also systematically discuss various aspects of LLM-based AI Agents \cite{wang2023survey, xi2023rise}.

\subsection{Task Planning}
LLMs are pre-trained on huge text corpora and present significant common sense reasoning and multi-task generalization abilities. Prompting is a highly effective method for further harnessing the intrinsic capabilities of LLMs to address various problems\cite{wei2022chain,kojima2022large}. For task planning, prompting facilitates LLMs to break down high-level tasks into sub-tasks\cite{huang2022language} and formulate grounded plans\cite{ahn2022can, huang2022inner}. ReAct\cite{yao2022react} proposes an enhanced integration of reasoning and action, enabling LLMs to provide a valid justification for action and integrating environmental feedback into the reasoning process. BabyAGI, AgentGPT, and AutoGPT also adopt step-by-step thinking, which iteratively generates the next task by using LLMs, providing some solutions for task automation.  However, these methods become problematic as an initial error can propagate along an action sequence, leading to a cascade of subsequent errors. Reflexion\cite{shinn2023reflexion} incorporates a mechanism for decision retraction, asking LLMs to reflect on previous failures to correct their decision-making. HuggingGPT\cite{shen2023hugginggpt} adopts a global planning strategy to obtain the entire sub-task queue within one user query. It is difficult to judge whether iterative or global planning is better since each one has its deficiencies and both of them heavily rely on the ability of LLMs, despite these models not being specifically tailored for task planning. Besides the above LLM-based studies, previous hierarchical agents, such as SEIHAI \cite{mao2022seihai}, Juewu-MC \cite{lin2021juewu}, GITM \cite{zhu2023ghost} often resemble the spirit of task planning.

However, in real-world systems, the high-level tasks are more intricate, and the prompting method without enhancing the intrinsic task-planning ability of LLMs can hardly achieve good performance. Thus, in our work, we adopt a fine-tuning mechanism to the planning dataset, along with well-designed prompts, to maximize the ability of task planning.

\subsection{Tool Usage}
The initial research in tool learning is limited by the capabilities of traditional deep learning approaches because of their weaknesses in comprehension of tool functionality and user intentions, as well as common sense reasoning abilities. Recently, the advancement of LLM has marked a pivotal juncture in the realm of tool learning. The great abilities of LLMs in common sense cognition and natural language processing attributes furnish indispensable prerequisites for LLMs to comprehend user intentions and effectively employ tools in tackling intricate tasks\cite{qin2023tool}. Additionally, tool usage can alleviate the inherent limitations of LLMs, encompassing the acquisition of up-to-date information from real-world events, enhanced mathematical computational abilities, and the mitigation of potential hallucinatory phenomena\cite{mialon2023augmented}.

In the domain of embodied intelligence\cite{duan2022survey}, LLMs directly interact with tangible tools, such as robots, to augment their cognitive abilities, optimize work productivity, and broaden functional capacities.LLM possesses the capability to automatically devise action steps according to user intentions, facilitating the guidance of robots in task completion\cite{zhang2023lp, shah2023lm, brohan2023can, huang2022inner, chen2023open, driess2023palm, wake2023chatgpt, rana2023sayplan, song2022llm}, or alternatively, to directly generate underlying code that can be executed by robots\cite{brohan2022rt, stone2023open, reed2022generalist, vemprala2023chatgpt, liang2023code}.

In addition to directly influencing the physical real world through interactions with tools, LLM can also utilize software tools such as search engines
\cite{guu2020retrieval, borgeaud2022improving}, mobile\cite{wang2023enabling, zhang2023mobile}, Microsoft Office \cite{li2023sheetcopilot, zha2023tablegpt}, calculators\cite{chen2023chatcot, parisi2022talm, cobbe2021training}, deep models\cite{gupta2023visual, chen2023language} and other versatile APIs\cite{lu2023chameleon,gou2023critic,liang2023taskmatrix} to improve model performance or complete complex workflows through flexible control of the software.

However, most of the aforementioned works focus only on specific scenarios, addressing how to choose or use the appropriate tools from a limited set, while agents in real-world scenarios usually have to face various and complex situations, requiring precise selection and usage of the correct tools from an API cloud with massive APIs. Gorilla\cite{patil2023gorilla} connects LLMs with massive APIs, which are, nonetheless, not real-world APIs and with poor diversity. ToolAlpaca\cite{tang2023toolalpaca} builds a tool-using corpus containing 3938 tool-use instances from more than 400 real-world tool APIs spanning 50 distinct categories, but this method focuses on smaller language models. ToolLLM\cite{qin2023toolllm} provides a novel and high-quality prompt-tuning dataset, ToolBench, which collects 16464 real-world APIs spanning 49 categories from RapidAPI Hub, covering both single-tool and multi-tool scenarios. TaskMatrix.AI\cite{liang2023taskmatrix} uses LLM as a core system and connects with millions of APIs to execute both digital and physical tasks. The methods above are of great assistance to the tool-learning research community.

 To augment LLMs with external tools, most recent methods rely on few-shot prompting with the off-the-shelf LLMs\cite{patil2023gorilla,tang2023toolalpaca,yao2023tree,wang2023voyager,li2023api, xu2023tool}
, but the existing LLMs are not developed for agentic use cases. FireAct\cite{chen2023fireact} proposes a novel approach to fine-tune LLMs with trajectories from multiple tasks and prompting methods and find LLM-based agents are consistently improved after fine-tuning their backbone. ToolLLM\cite{qin2023toolllm} uses SFT based on the proposed ToolBench, to transform LLaMa\cite{touvron2023llama} into ToolLLaMa, which demonstrates a remarkable ability to execute complex instructions and generalize to unseen APIs, and exhibits comparable performance to ChatGPT.
Inspired by these, we not only design an API Retriever and Demo Selector to serve as an auto-prompter but also employ fine-tuning techniques to further enhance the performance of our framework so that it can address much more complex tasks in real-world scenarios.

\section{Conclusion}

In this paper, we present a comprehensive framework designed to augment the capabilities of Large Language Models (LLMs) in complex, real-world scenarios, particularly focusing on task planning and tool usage. Our approach, which integrates the API Retriever, LLM Finetuner, and Demo Selector, has been rigorously tested and validated in various settings. 
The results demonstrate that fine-tuning LLMs with a curated dataset significantly improves their effectiveness in executing real-world tasks. The API Retriever and Demo Selector components also prove indispensable, particularly in enhancing the model's decision-making accuracy and adaptability. This research not only showcases the potential of LLMs in practical applications but also lays a foundation for future advancements in the field. 
By addressing the challenges of API diversity and complexity, our framework paves the way for more efficient, and user-centric AI systems, capable of handling real-world scenarios.

\section*{Acknowledgements}

This work was conducted collaboratively among the authors.

Hangyu Mao and Rui Zhao led the project. 

Regarding the implementation and evaluation phase, Yihong Chen, Tianpeng Bao, Guoqing Du, Xiaoru Hu, Shiwei Shi, Jingqing Ruan, Yilun Kong and Bin Zhang performed the experiments and analyzed the data. Hangyu Mao assisted in the analysis of the experimental phenomena and offered constructive suggestions for improvements. Ziyue Li, Xingyu Zeng and Rui Zhao provided invaluable feedback, contributed to the direction of the research. All authors participated in the discussion.

Regarding the manuscript phase, Jingqing Ruan and Yilun Kong organized and wrote main parts of this manuscript. Hangyu Mao provided assistance during the process.  Each author read and approved the final manuscript. 

The authors would like to thank Feng Zhu, Kun Wang, Yuhang Ran, and colleagues from the product-side for their valuable feedback, discussion, and participation in this project.

\bibliographystyle{IEEEtran}
\bibliography{ref}

\end{document}